# An Integrated System for 3D Gaze Recovery and Semantic Analysis of Human Attention

Lucas Paletta, Katrin Santner and Gerald Fritz

JOANNEUM RESEARCH Forschungsgesellschaft mbH
DIGITAL - Institute for Information and Communication Technologies
Steyrergasse 17, Graz, Austria
`lucas.paletta@joanneum.at`

**Abstract.** This work describes a computer vision system that enables pervasive mapping and monitoring of human attention. The key contribution is that our methodology enables full 3D recovery of the gaze pointer, human view frustum and associated human centered measurements directly into an automatically computed 3D model in real-time. We apply RGB-D SLAM and descriptor matching methodologies for the 3D modeling, localization and fully automated annotation of ROIs (regions of interest) within the acquired 3D model. This innovative methodology will open new avenues for attention studies in real world environments, bringing new potential into automated processing for human factors technologies.

## 1    Introduction

This work presents a computer vision system methodology that, *firstly*, enables to precisely estimate the position and orientation of human view frustum and gaze and from this enables to precisely analyze human attention in the context of the semantics of the local environment (objects, signs, scenes, etc.). Figure 1 visualizes how accurately human gaze is mapped into the 3D model for further analysis. *Secondly*, the work describes how ROIs (regions of interest) are automatically mapped from a reference video into the model and from this prevents from state-of-the-art laborious manual labeling of tens / hundreds of hours of eye tracking video data. This will provide a scaling up of nowadays still small sketched usability studies and thus enable for the first time statistically significant usability engineering studies.

The methodology for the recovery of human attention in 3D environments is based on the workflow as sketched in Figure 2: For a spatio-temporal analysis of human attention in the 3D environment, we firstly build a spatial reference in terms of a three-dimensional model of the environment using RGB-D SLAM methodology. The user's view is then gathered with eye tracking glasses (ETG) and localized within the 3D model. ROIs are marked on imagery and automatically detected in video and then mapped into the 3D model. The distribution of saliency onto the 3D environment is computed for further human attention analysis. The performance evaluation of the presented methodology refers to results from a dedicated test environment (Paletta et al. [1]) where we demonstrate very low angular projection errors ($\approx 0.6^{\circ}$, $\approx 1.1$ cm) which will enable to capture attention on daily objects and activities (package logos, cups, books, pencils).

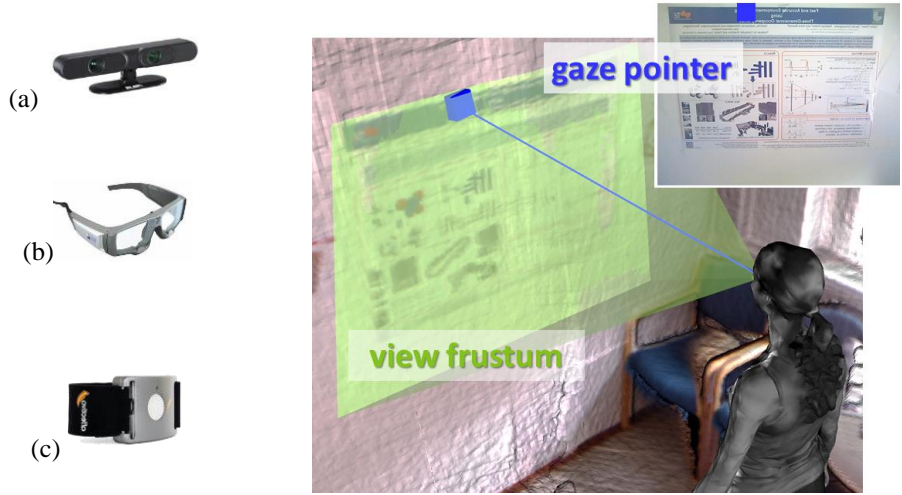

**Figure 1.** Sketch of sensors used in the study (left) and typical gaze recovery (right). A full 6D recovery of the view frustum and gaze (right) is continuously mapped into the 3D model. (a) RGB-D scanning device, (b) eye tracking glasses (ETG) and (c) bio-electrical signal device.

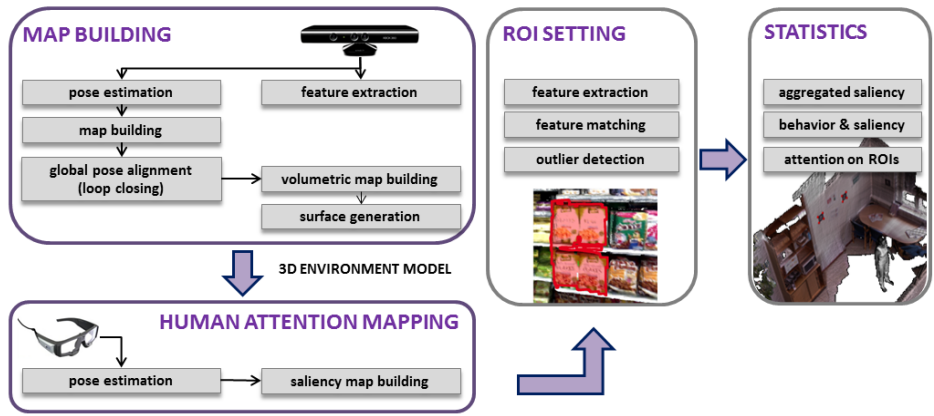

**Figure 2.** Sketch of workflow for 3D gaze recovery and semantic ROI analytics.

## 2    Related Work

**Human Attention Analysis in 3D**. 3D information recovery of human gaze has in principle been targeted by Munn et al. [2] who introduced monocular eye-tracking and triangulation of 2D gaze positions of subsequent key video frames, obtaining observer position and gaze pointer in 3D with angular errors of ≈3.8°. Pirri et al. [3] achieved accuracy indoors about ≈3.6 cm at 2 m distance to the target compared to our ≈0.9 cm [1]. Summarizing, previous attempts only covered several aspects of the problem while we present a straight forward solution of mapping fixation distributions onto saliency maps within a model of the environment, and the possibility of real-time tracking of attention with mass marketed eye-tracking hardware.

## 3   Gaze Localization in 3D Models

**Visual Map Building and Camera Pose Estimation**. For realistic environment modeling we make use of an RGB-D sensor providing per pixel color and depth information at high frame rates. Our environment consists of a sparse pointcloud, where each landmark [4] is attached for data association during pose tracking. Estimated camera poses are stored in a 6DOF manner. Incremental camera pose tracking assuming an already existing map is done by keypoint matching followed by a least-square optimization routine minimizing the reprojection error of 2D-3D correspondences.

**Densely Textured Surface Generation**. For realistic environment visualization, user interaction and subsequent human attention analysis, a dense, textured model of the environment has to be constructed. Therefore, depth images are integrated into a 3D occupancy grid [5] using the previously corrected camera pose estimates.

**3D Gaze Recovery from Monocular Localization.** To estimate the proband's pose, SIFT keypoints are extracted from ETG video frames and a *full 6DOF pose* is estimated using the perspective n-Point algorithm [6].

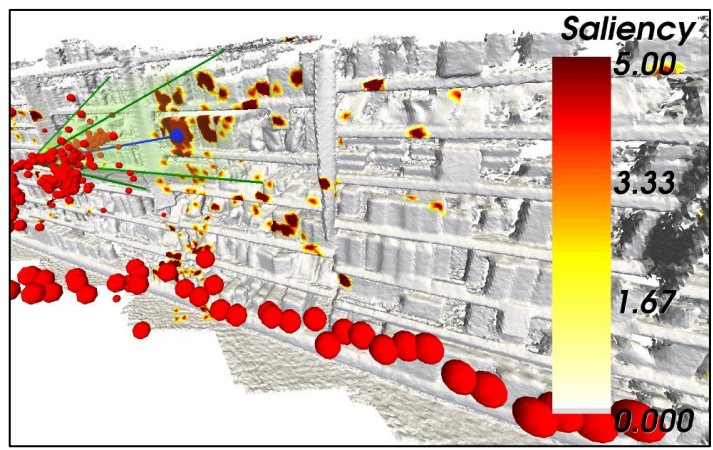

**Figure 3.** Mapping of saliency onto the acquired 3D model and automated recovery of the trajectory of ETG camera positions (spheres). Recovery of frustum and gaze.

**Automated 3D Annotation of Regions of Interest.** Annotation of ROIs in 2D or even 3D information usually causes a process of massive manual interaction. In order to map objects of interests, such as, logos, package covers, etc. into the 3D model, we first use logo detection in the high resolution scanning video to search for occurrences of predefined reference appearances, using vocabulary trees [4].

**Semantic Mapping of Attention.** The automatic detection of ROIs in 3D enables the system to provide the user with statistical evaluation, such as on ROIs called AOI hit, which states for a raw sample or a fixation that its coordinate value is inside the ROI [7]. Figure 6 plots the distribution of the dwell times for ROI #1 over all participants, we can conclude that some of the captured fixations is related to human object recognition which is known to trigger from 100 ms of observation / fixation [8].

## 4  Experimental results

**Eye Tracking Device.** The mass marketed SMI™ eye-tracking glasses (Figure 1b) - a non-invasive video based binocular eye tracker with automatic parallax compensation - measures the gaze pointer for both eyes with 30 Hz. The gaze pointer accuracy of 0.5°–1.0° and a tracking range of 80°/60° horizontal/vertical assure a precise localization of the human's gaze in the HD 1280x960 scene video with 24fps. We recorded data on a shop floor covering an area of about 8x20m². We captured 2366 RGB-D images and reconstructed the environment consisting of 41700 natural visual landmarks and 608 keyframes.

## 5  Conclusion and Future Work

We present a complete system for (i) wearable data capturing, (ii) automated 3D modeling, (iii) automated recovery of human pose and gaze, and (iv) automated ROI based semantic interpretation of human attention. The presented system is a significant step towards a mobile mapping framework [9] for quantitative analysis of human attention information [10] in natural environments. Future work will focus on improved tracking of the human pose across image blur and uncharted areas as well as study human factors in the frame of stress and emotion in the context of the 3D space.

### Acknowledgments

This work has been partly funded by the European Community's FP7/2007-2013, grant agreement n°288587 MASELTOV, and by the Austrian FFG, contract n°832045, Research Studio Austria FACTS.

### References

1. Paletta, L., Santner, K., Fritz, G., Mayer, H., & Schrammel, J.: 3D Attention: Measurement of Visual Saliency Using Eye Tracking Glasses, *Proc. CHI 2013*.
2. Munn, S. M., & Pelz J. B.: 3D point-of-regard, position and head orientation from a portable monocular video-based eye tracker. *Proc. ETRA 2008,* pp. 181-188.
3. Pirri, F., Pizzoli, M., & Rudi, A.: A general method for the point of regard estimation in 3D space. *Proc. CVPR 2011*, pp. 921-928.
4. Nistér, D. & Stewénius, H.: Scalable Recogn. with a Vocabulary Tree, *Proc. CVPR 2006.*
5. Marks, T. K., Howard, A., Bajracharya, M., Cottrell, G. W. & Matthies, L.: Gamma-SLAM: Using stereo vision and variance grid maps for SLAM, *Proc. ICRA 2008*.
6. Lepetit V., Moreno-Noguer F. and Fua P.: EPnP: An Accurate O(n) Solution to the PnP Problem, *International Journal of Computer Vision*, pp. 155-166, 2009.
7. Holmqvist, K., Nyström, M., Andersson, R., Dewhurst, R., Jarodzka, H., and van de Weijler, J.: Eye Tracking, Oxford University Press, 2011, pp. 187.
8. Grill-Spector, K. and Sayres, R. Object Recognition: Insights From Advances in fMRI Methods, *Current Direct. in Psychol. Science*, Vol. 17, No. 2. 2008, pp. 73-79.
9. Paletta, L., Santner, K., Fritz, G., Hofmann, A., Lodron, G., Thallinger, G., and Mayer, H. FACTS - A Computer Vision System for 3D Recovery and Semantic Mapping of Human Factors, *Proc. ICVS 2013*, Springer-Verlag, LNCS 7963, pp. 62-72.
10. Fritz, G., Seifert, C., Paletta, L., and Bischof, H. (2005). Attentive object detection using an inform. theoret. saliency measure, *Proc. WAPCV 2004,* Springer-Verlag, LNCS 3368.